\documentclass[12pt]{article}

\usepackage{amssymb}
\usepackage{amsmath}
\usepackage{hyperref}  %% ensures inline citations link to references
\hypersetup{pdfborder=0 0 0} %% suppress borders around hyperlinks
\hypersetup{  %% set up consistent link colours
    colorlinks=true,
    linkcolor=blue,
    citecolor=blue,
    filecolor=magenta,      
    urlcolor=blue,
    pdfstartview=FitH,
    pdftitle={Stable Structure Learning with HC-Stable and Tabu-Stable Algorithms.},
    pdfauthor={N. K. Kitson and A. Constantinou},
    pdfsubject={Structure Learning, instability},
    pdfkeywords={Bayesian Networks, algorithm stability, hill-climbing, Tabu algorithm}}

\usepackage{dirtytalk}  %% used for quoted text
\usepackage[nofillcomment,ruled]{algorithm2e}  %% used for algorithm pseudocode

\usepackage{booktabs}  %% nicer styling for tables without vertical lines
\usepackage{makecell}
\usepackage{pifont}  %% supports "cross" symbols
\newcommand{\rx}{\color{red}\ding{54}}  %% red cross symbol
\newcommand{\bx}{\color{blue}\ding{54}}  %% blue cross symbol
\usepackage{float}  %% more control over the positioning of floats
\usepackage{colortbl}  %% for alternating coloured table rows
\definecolor{palegrey}{gray}{0.93}  %% pale grey for table rows
\newcolumntype{R}[1]{>{\raggedleft\arraybackslash}p{#1}}  %% custom right-justified, wrapped, column type
\newcommand{\B}[1]{\textbf{#1}}  %% custom bolded table value
\newcommand{\W}[1]{\textbf{\textcolor{red}{#1}}}  %% custom bolded, red, table value

\usepackage[numbers]{natbib} %% not needed if have bbl file
\usepackage{graphicx}

\usepackage{authblk}  %% layout author and affiliations nicely

\title{Stable Structure Learning with HC-Stable and Tabu-Stable Algorithms.}

\author{Neville K. Kitson\thanks{\texttt{n.k.kitson@qmul.ac.uk} (corresponding author)} }
\author{Anthony C. Constantinou\thanks{\texttt{a.constantinou@qmul.ac.uk}}}
\affil{Bayesian AI research lab,\\
       Machine Intelligence and Decision Systems (MInDS) group, \\
       School of Electronic Engineering and Computer Science, \\
       Queen Mary University of London (QMUL)}

\date{\small}

\begin{document}

\vspace{-1cm}
\maketitle

%% \author{Neville K. Kitson \corref{cor1}}
%% \ead{n.k.kitson@qmul.ac.uk}
%% \cortext[cor1]{Corresponding author}
%% \author{Anthony C. Constantinou}
%% \ead{a.constantinou@qmul.ac.uk}


% \author{
%    Neville K. Kitson and Anthony C. Constantinou \\
%     Machine Intelligence and Decision Systems (MInDS) group, \\
%     Queen Mary University of London (QMUL) \\
%     \texttt{n.k.kitson@qmul.ac.uk, a.constantinou@qmul.ac.uk}
% }

%% \affiliation{organization={Machine Intelligence and Decision Systems (MInDS) group,
%%                           Queen Mary University of London (QMUL)},%Department and Organization
%%            city={London},
%%             postcode={E1 4NS}, 
%%            country={United Kingdom}}

\vspace{-1cm}  % Adjust this value as needed
\begin{abstract}
Many Bayesian Network structure learning algorithms are unstable, with the learned graph sensitive to arbitrary dataset artifacts, such as the ordering of columns (i.e., variable order). PC-Stable \cite{colombo2014order} attempts to address this issue for the widely-used PC algorithm, prompting researchers to use the ‘stable’ version instead. However, this problem seems to have been overlooked for score-based algorithms. In this study, we show that some widely-used score-based algorithms, as well as hybrid and constraint-based algorithms, including PC-Stable, suffer from the same issue. We propose a novel solution for score-based greedy hill-climbing that eliminates instability by determining a stable node order, leading to consistent results regardless of variable ordering. Two implementations, HC-Stable and Tabu-Stable, are introduced. Tabu-Stable achieves the highest BIC scores across all networks, and the highest accuracy for categorical networks. These results highlight the importance of addressing instability in structure learning and provide a robust and practical approach for future applications. This extends the scope and impact of our previous work presented at Probabilistic Graphical Models 2024 \citep{kitson2024eliminating} by incorporating continuous variables. The implementation, along with usage instructions, is freely available on GitHub at \url{https://github.com/causal-iq/discovery}.
\vspace{1cm}
\end{abstract}

%% \end{frontmatter}


\section{Introduction}
\label{sec:intro}

Bayesian Networks (BNs) \citep{koller2009probabilistic} are an approach for modelling complex probabilistic relationships in diverse domains such as healthcare \citep{kyrimi2021comprehensive}, fault diagnosis \citep{cai2017bayesian} and the environment \citep{vitolo2018modeling}. They can be used to answer \textit{probabilistic queries} which predict the probability distribution of a subset of variables conditional on the values of other variables, and so can answer questions such as \textit{if these symptoms are present, what is the probability the patient has disease Y?}. Moreover, if one additionally assumes that the relationships are \textbf{\textit{causal}}, then the resulting \textit{Causal} BN can be used to answer \textit{interventional queries} such as \textit{if the patient is given this treatment, what is the likely outcome?} \citep{pearl2009causality}. Thus, BN's have an important potential role as A.I. decision support systems.

Because BNs are probabilistic graphical models, one key challenge is to specify the graphical structure underlying them. Using machine learning to infer this structure from observational data is an active research area. Recent work in \cite{kitson2024impact} shows that many algorithms are \textbf{\textit{unstable}}; that is, sensitive to artifacts of the data, such as the ordering of columns in the data. This paper shows that this instability can affect the objective score and the structural accuracy of the learned graph, as well as the time taken to learn the graph. It extends the scope and impact of our previous work presented at Probabilistic Graphical Models 2024 \citep{kitson2024eliminating} by incorporating continuous variables and extending the analysis to consider the effect of instability on other properties of the learned graph such as objective score and density.

The principal contribution of this paper is to describe an approach which eliminates this instability for greedy score-based hill-climbing algorithms. We show that one of our implementations, Tabu-Stable, is completely stable \textbf{\textit{and}} learns higher-scoring categorical and continuous variable graphs than all of the well-known algorithms that we compare it to. Tabu-Stable also learns the most structurally accurate graphs when learning categorical networks. The approach described in this paper also has wider applicability to many algorithms that make use of greedy hill-climbing, such as hybrid algorithms.

\section{Background}
\label{sec:background}

\subsection{Bayesian Networks}
\label{sub:bayesian_networks}

The key element of a Bayesian Network is a Directed Acyclic Graph (DAG) where each node represents a variable and the directed edges, or \textit{arcs}, represent a dependence relationship between the two variables. We denote the $n$ variables in the BN as $X_1, ..., X_n$. If there is an arc $X_A \longrightarrow X_B$, $X_A$ is termed the \textit{parent} of $X_B$. The DAG is constructed so that a node is conditionally independent of all variables except its descendants given its parents. This is called the \textit{Local Markov Property} and allows the global probability distribution to be expressed compactly as:
\begin{equation}
P(X_1, X_2, ..., X_n)=\prod_{i=1}^n P(X_i | \textbf{Pa}(X_i))
\end{equation}
where $\textbf{Pa}(X_i)$ are the parents of $X_i$.

A key result flowing from the Local Markov Property is that a graphical property of the DAG, \textit{d-separation}, is equivalent to variables being conditionally independent of one another \citep{pearl1988probabilistic}. D-separation can be used to infer the graph structure from the independence and dependence relationships present in the data. In general, however, more than one DAG is consistent with the independence relationships \citep{verma1990equivalence}. These \textit{Markov Equivalent} graphs belong to a \textit{Markov Equivalent Class (MEC)}. A MEC is usually represented by a \textit{Completed Partially Directed Acyclic Graph (CPDAG)}, where directed edges indicate edges where all DAGs in the MEC have that same orientation, and undirected edges indicate that some DAGs have one orientation and the rest the other. 

The other component of a BN is the specification of the probabilistic dependence relationship between adjacent variables and the probability distributions assumed. For the categorical variable networks considered here, this takes the form of \textit{Conditional Probability Tables (CPTs)} which define a multinomial distribution for the child values for each combination of parental values. For the continuous variable networks  considered, it is assumed that the value of a child variable is a linear function of the values of its parents plus a Gaussian noise term. These are termed \textit{Gaussian Bayesian Networks} and the resulting global probability distribution is a multivariate Gaussian.

\subsection{Structure Learning Algorithms}
\label{sub:structure_learning}

The specification of a BN's DAG structure may be undertaken using human expertise, using a structure learning algorithm to learn it from data, or a combination of both. Structure learning algorithms usually learn from observational data, since it is more readily available. Algorithms typically make assumptions that often do not hold in practice. For example, that there are no missing data, measurement error, or latent confounders. 

\textit{Constraint-based} algorithms such as PC \citep{spirtes1991algorithm}, GS \citep{margaritis1999bayesian} and Inter-IAMB \citep{tsamardinos2003time} use statistical conditional independence (CI) tests to identify the independence relationships in the data, and use the d-separation principle to infer the DAG structure. Constraint-based algorithms usually assume \textit{faithfulness}, which states that there are no independence relationships in the data which are not implied by the DAG. Variants of the approach such as FCI \citep{spirtes2001causation} can account for latent confounders. Most constraint-based algorithms aim to return a CPDAG which represents all the possible DAGs that are consistent with the independence and dependence relationships in the data. However, errors arising from the unreliability of the independence tests used or algorithm design may mean that they sometimes return a graph containing directed and undirected edges which is not consistent with any possible DAG. In that case, the learned graph is termed a \textit{non-extendable Partial DAG (PDAG)}.

\textit{Score-based} algorithms represent the second major class of algorithms. These follow a more traditional machine-learning approach of using an objective function to assign a score to each graph and then employ some strategy to find a high-scoring graph. The Hill-Climbing (HC) \citep{herskovits1990kutato} and Tabu \citep{bouckaert1995bayesian} algorithms are two simple score-based algorithms that remain competitive and commonly-used. Other score-based approaches search through MEC space, for example, GES  \citep{chickering2002optimal} and FGES \citep{ramsey2017million}. \textit{Exact} score-based algorithms, such as GOBNILP \citep{bartlett2017integer} guarantee to return the highest scoring graph, though typically with limits placed on the number of parents of any node.

Other classes of algorithms include \textit{hybrid} ones such as MMHC \citep{tsamardinos2006max} and H2PC \citep{gasse2014hybrid} which use a mix of score and constraint-based methods. More recent developments include algorithms which make additional assumptions about the functional relationships between variables to identify arc orientations \citep{peters2014causal}, and algorithms such as NOTEARS \citep{zheng2018dags} which treat structure learning as a continuous optimisation problem. \cite{kitson2023survey} provides a comprehensive survey across the different classes of algorithms.

The objective function in score-based approaches usually includes the log-likelihood of the data being generated from the graph, with two commonly-used scores being BIC \citep{suzuki1999learning} and BDeu \citep{heckerman1995learning}. The BIC score, $S_{BIC}$, for a graph $G$ with $n$ variables and dataset $D$ is computed as follows:
\begin{equation} \label{eqn:bic}
S_{BIC}(G, D) = \sum_{i=1}^{n} \sum_{j=1}^{q_i} \sum_{k=1}^{r_i}
\left[ N_{ijk} \log{\frac {N_{ijk}} {N_{ij}}} \right]
- \frac{\log{N}}{2} \cdot F
\end{equation}
The first term on the right-hand side of Equation~\ref{eqn:bic} is the log-likelihood and is based on counts of values in the dataset. Specifically, $N_{ijk}$ is the number of rows where node $X_i$ has the $k^{th}$ out of the $r_i$ possible values and its parents $\textbf{Pa}(X_i)$ have the $j^{th}$ combination of values of the $q_i$ possible combinations, and $N_{ij}$ is the total number of rows where the parents have that $j^{th}$ combination of values. The second term is a model complexity penalty, where $N$ is the total number of rows in $D$, and $F$ is the number of free parameters in the CPTs of the model.

BDeu, $S_{BDeu}$, is a Bayesian score representing the posterior probability of graph $G$ given the data $D$ assuming some prior beliefs about the probability of each graph and set of parameter values. If all graphs are assumed equally probable, then BDeu is given by:
\begin{equation} \label{eqn:bdeu}
S_{BDeu}(G, D) = \sum_{i=1}^{n} \sum_{j=1}^{q_i} 
\left[ \log{\frac{\Gamma(\frac{N'}{q_i})}{\Gamma(N_{ij} + \frac{N'}{q_i})}} 
+ \sum_{k=1}^{r_i} \log{\frac{\Gamma(N_{ijk} + \frac{N'}{r_i q_i})} {\Gamma(\frac{N'}{r_i q_i})}} \right]
\end{equation}
where $\Gamma$ is the Gamma function, $N'$ assigns a weight to the prior parameter beliefs, and other symbols take the same meanings as in Equation~\ref{eqn:bic}. BIC and BDeu are \textit{decomposable} scores, since they represent the sum of individual scores for each node. This property facilitates the efficient re-computation of the graph score as the graph is modified. Both are also \textit{score equivalent} which means they assign the same score to all DAGs in a particular MEC.

\subsection{Related Work}
\label{sub:related_work}

The instability of structure learning algorithms, and specifically their sensitivity to arbitrary dataset artifacts such as \textit{variable order}, seems to have attracted relatively little attention. Variable order in this paper means the order of the columns in the dataset, something which is arbitrary and ideally ought to have no effect on the learned graph. Exact algorithms such as A-Star \citep{yuan2011learning} which implicitly determines the best node order, and those that search over MECs such as GES and FGES should, \textbf{\textit{in principle}}, be insensitive to dataset artifacts. Algorithms that rely on a topological order being specified such as K2 \citep{cooper1992bayesian}, or ones that generate an order themselves \citep{larranaga1996learning,behjati2020improved}, should also be stable. Approaches that average over several learned graphs, either those sampled from a posterior distribution of graphs such as Order-MCMC \citep{friedman2003being}, or which use different sub-samples of data \citep{broom2012model} or different classes of learner \citep{constantinou2023open} might be less sensitive since any effect of dataset artifacts may tend to 'cancel out' over the population of learned graphs.

Nonetheless, the stability of algorithms is rarely explicitly considered or evaluated. An exception is PC-Stable \citep{colombo2014order} where the authors strive to minimise the effect of node processing order in the PC algorithm which they find has a considerable effect on how errors propagate throughout the learning process. PC-Stable offers better accuracy and lower sensitivity to variable order than PC, and it is generally chosen over PC for that reason. However the results in Subsection~\ref{sub:other_algorithms} demonstrates that it retains a considerable amount of instability. \cite{kitson2024impact} shows that variable order can impact the ranking of algorithms, but it is not usually considered when comparing algorithms. \cite{scutari2019learns} tries a small number of different orderings in order to improve arc orientations as part of a comparative benchmark, but sensitivity to ordering is not reported.

\section{Eliminating Instability in Hill-Climbing}
\label{sec:eliminating-instability}

This section discusses the source of instability within hill-climbing algorithms and how it can be addressed. The focus is on the Tabu algorithm since it is commonly used and is competitive in benchmarks, but we apply the same approach to the simpler HC algorithm. The resulting two new algorithms, Tabu-Stable and HC-Stable, are described.

\subsection{How Instability Arises in Hill-Climbing Algorithms}
\label{sub:instability}

HC is a greedy, score-based hill-climbing algorithm. It typically starts exploring the search-space of graphs from an empty DAG, and searches for the single arc addition, deletion or reversal which increases the objective score of the DAG the most at each iteration. Changes that would create a cycle are not considered. The algorithm terminates when there are no further changes that would increase the score, and the resulting DAG generally has only a locally-maximum score. 

Tabu is a higher-performing variant of HC that allows iterations where the score stays the same or decreases allowing the algorithm to escape some local maxima. Tabu maintains a fixed-length list of recently visited DAGs, \textit{tabulist}, to prevent the algorithm from repeatedly considering previously-visited DAGs. The black-coloured pseudo-code in Algorithm~\ref{algo:tabu-stable} shows the main elements of Tabu. The $deltas$ variable holds the score change associated with every possible change to the DAG. Lines 5 to 25 form the main iterative loop, with the highest-scoring change for each iteration identified in the \textbf{foreach} loop, and applied to the DAG at line 19. $UpdateDeltas$ updates $deltas$ appropriately following the change; for example, adding arc $A \longrightarrow B$ would mean that deltas for adding arcs which point to $B$ must be recalculated to take into account that $B$ now has a new parent. The $stop\_condition$ for the main loop is that none of the last $noinc$ (a hyperparameter) changes have increased the score. HC is similar to Algorithm~\ref{algo:tabu-stable} except that $tabulist$ is not required, and the $stop\_condition$ is that there are no further changes which will \textit{\textbf{increase}} the score.

\begin{figure}[ht]
    \centering
    \includegraphics[width=1.0\linewidth]{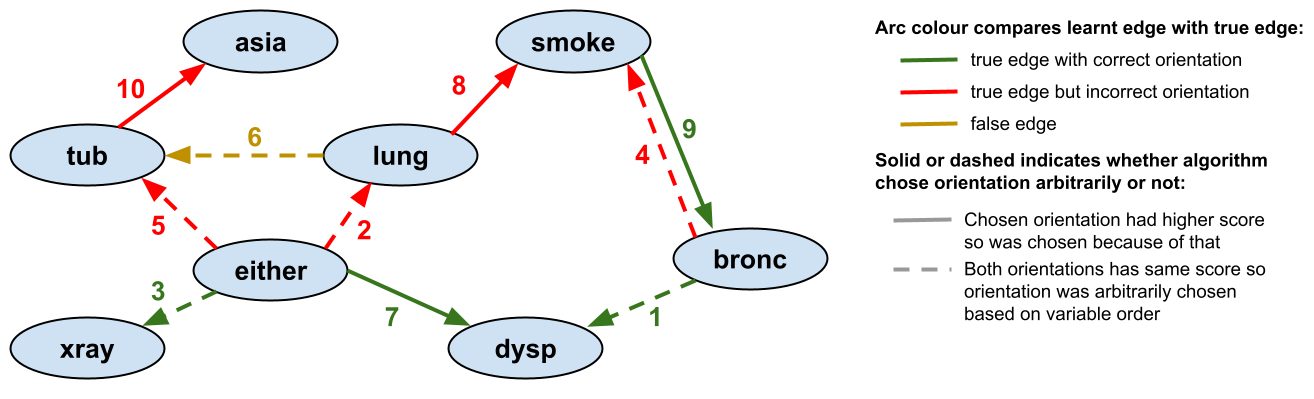}
    \caption{The sequence of DAG changes when HC learns the categorical Asia network from 10,000 samples. The numbers beside each arc show the iteration at which it is added. Arc colours compare the learned arc against the true arc, and whether it is solid or dashed indicates whether its orientation was arbitrary. Variable order within the dataset is alphabetic.}
    \label{fig:arc_orientation}
\end{figure}

Figure \ref{fig:arc_orientation} illustrates the source of instability in hill-climbing by showing the sequence in which HC conventionally learns the Asia network from $10^4$ rows of data, when using the BIC score as the objective function. Both orientations of the edge $bronc \textbf{\textemdash} dysp$ give the same maximal score improvement at the first iteration. If HC is implemented to use the variable order to orientate the arc in this situation, and that order is alphabetic, then orientation $bronc \longrightarrow dysp$ would be chosen. This \textit{\textbf{arbitrary}} orientation happens to agree with the true graph. The orientation of the second arc added, $either \rightarrow lung$, is similarly arbitrary, but in this case, incorrect.  Just like in the case of constraint-based learning with the PC algorithm, \textit{\textbf{this instability propagates to subsequent iterations}}. For example, if the variable order had been such that the second iteration correctly added $lung \rightarrow either$, then edge $tub \textbf{\textemdash} either$ would also have been orientated correctly. This, in turn, would have stopped the extraneous arc $lung \rightarrow tub$ being added. While the impact of variable order varies from network to network, \cite{kitson2024impact} shows that it is generally considerable, typically overshadowing the impact of changing objective functions, sample size or hyperparameters.

\subsection{Stabilising Hill-Climbing}
\label{sub:stabilising-tabu}

Algorithm~\ref{algo:tabu-stable} illustrates the key elements, shown in red pseudocode, of the new algorithm Tabu-Stable that avoids arbitrary orientations and hence becomes completely insensitive to the ordering of the variables as read from data. The key is to determine a \textit{stable order} at line 4 that is not dependent on the artifacts of the dataset, such as the variable order. The DAG change with the highest score improvement at each iteration is determined in the \textbf{for each} loop as usual, but this now also records whether there is an equivalent change ($equiv\_change$), which adds an arc in the opposite orientation with the same maximum score improvement. In that case, the one consistent with $stable\_order$ is added to the DAG. Analogously to the relationship between Tabu and HC, HC-Stable is simply Tabu-Stable with $tabulist$ removed and a $stop\_condition$ that there are no further changes that would \textbf{\textit{increase}} the score.

\begin{algorithm}[H]
\caption{Tabu-Stable (changes to standard Tabu shown in red)}
\label{algo:tabu-stable}

\DontPrintSemicolon
\LinesNumbered

\KwIn{$data$, dataset to learn graph from}
\KwOut{$best\_dag$, highest-scoring DAG found}

\BlankLine

\textcolor{red}{$stable\_order \leftarrow GetStableOrder(data)$} 
(see Algorithm~\ref{algo:get-stable-order}) \;
{$best\_dag \leftarrow dag \leftarrow$ empty DAG } \;
{$tabulist \leftarrow$ empty list } \;
{$deltas \leftarrow$ score change for each arc addition} \;

\BlankLine

\Repeat{stop\_condition}{
    \BlankLine
    $max\_delta \leftarrow None$ \;
    \ForEach{$dag\_change = AllowedChange(dag, tabulist)$}{
        \uIf{max\_delta = None or $delta[dag\_change] > max\_delta$}{
            $max\_delta \leftarrow delta[dag\_change]$ \;
            $best\_change \leftarrow dag\_change$ \;
            \textcolor{red}{$equiv\_change \leftarrow None$} \;
        }

\SetKwIF{If}{ElseIf}{Else}{\textcolor{red}{if}}{\textcolor{red}{then}}{\textcolor{red}{else if}}{\textcolor{red}{else}}{\textcolor{red}{end if}}%

        \ElseIf{\textcolor{red}{$AddingSameEdgeWithSameDelta(dag\_change, best\_change)$}}{
            \textcolor{red}{$equiv\_change \leftarrow dag\_change$} \;
        }
    }
    \BlankLine
  
    \SetKwIF{If}{ElseIf}{Else}{\textcolor{red}{if}}{\textcolor{red}{then}}{\textcolor{red}{else if}}{\textcolor{red}{else}}{\textcolor{red}{end if}}%
    \If{\textcolor{red}{$equiv\_change \neq None$ {\normalfont and} $equiv\_change$ consistent with $stable\_order$}}{
        \textcolor{red}{$best\_change \leftarrow equiv\_change$} \;
    }
    \SetKwIF{If}{ElseIf}{Else}{\textcolor{black}{if}}{\textcolor{black}{then}}{\textcolor{black}{else if}}{\textcolor{black}{else}}{\textcolor{black}{end if}}%
    
    \BlankLine
    $dag \leftarrow dag + best\_change$ \;
    $UpdateDeltas(deltas, best\_change)$ \;
    insert $dag$ into $tabulist$ \;
    \If{Score(dag, data) $>$ Score(best\_dag, data)}{
        $best\_dag \leftarrow dag$
    }
\BlankLine
}
\end{algorithm}
\vspace{1em}

The function \textit{GetStableOrder} shown in Algorithm~\ref{algo:get-stable-order} returns the stable node order used to avoid arbitrary orientations. This order is generated in two stages. Firstly, lines 1-11 of Algorithm~\ref{algo:get-stable-order} produce $dec\_score\_order$, which contains nodes primarily ordered by the decomposable objective score, BIC or BDeu for example, that will later be used in the structure learning itself. The sort key for the list has three elements, which in order of precedence are: (1) the score of the node without parents, (2) the mean score of the node where every other node is taken as its single parent (computed in lines 4 to 7), and (3), in the case of categorical variables only, a textual rendition of the counts of values of the variable, e.g. "\{'no': 5, 'yes': 3\}". The first two elements of the sort key use scores such as those described in Equation~\ref{eqn:bic} and Equation~\ref{eqn:bdeu} and are therefore determined solely by the \textbf{\textit{data distribution itself}}.

Moreover, since the second element of the sort key for a node compares the distribution of combinations of values of that node and every other node, the intuition is that it is very unlikely that two variables will have the same sort key unless they are indeed identical (duplicate). One situation where the first two elements of the sort key are the same for non-identical categorical variables is where the sequences of values for them are isomorphic, e.g. $a, a, a, c, c, a$ and $c, c, c, b, b, c$, but here the third element of the sort key will differ. If two variables do have the same sort key, then they revert to being ordered by variable order. This will most likely occur because the variables have identical sequences of values; that is, they are effectively duplicate variables. We find that this can occur learning from categorical variables with limited sample sizes. Thus, the whole algorithm retains some unavoidable sensitivity to variable order when there are identical (or duplicate) variables. This is explored in Subsection~\ref{sub:residual_instability}.

\begin{algorithm}[H]
\caption{GetStableOrder - determines a stable processing order}
\label{algo:get-stable-order}

\DontPrintSemicolon
\LinesNumbered

\SetKwFunction{FMain}{GetStableOrder}
\SetKwProg{Fn}{Function}{:}{}
\Fn{\FMain{$data$}}{

\KwIn{$data$, dataset learning graph from}
\KwOut{$stable\_order$, stable order for use in Tabu-Stable}

{$dec\_score\_order \leftarrow empty\_list$} \;
\ForEach{variable \textnormal{in} data}{
    $uncond\_score = NodeScore(variable)$ \;
    $cond\_score = 0.0$ \;
    \ForEach{\textnormal{possible} single\_parent \textnormal{of} variable \textnormal{in} data}{
        $cond\_score \leftarrow cond\_score + NodeScore(variable, single\_parent) / (n - 1)$  \;
    }
    $sorted\_value\_counts \leftarrow \textnormal{counts of unique values of } variable \textnormal{ in } data$ \;
    $sort\_key = (uncond\_score, cond\_score, sorted\_value\_counts)$ \;
    $dec\_score\_order \leftarrow InsertByKey(variable, sort\_key)$ \;
}
{$inc\_score\_order \leftarrow reverse(dec\_score\_order)$} \;
{$inc\_dag \leftarrow HC(data, inc\_score\_order)$} \;
{$dec\_dag \leftarrow HC(data, dec\_score\_order)$} \;
\uIf{$DAGScore(inc\_dag, data) > DAGScore(dec\_dag, data)$}{
    {$stable\_order = TopologicalOrder(inc\_dag)$}
}
\Else {
    {$stable\_order = TopologicalOrder(dec\_dag)$}
}
{\textbf{return}  $stable\_order$ } \;

}
\end{algorithm}
\vspace{1em}

It is not expected that $dec\_score\_order$ will necessarily be a \textit{\textbf{good}} order, rather, only that it will be insensitive to artifacts of the dataset such as row or column order. \cite{kitson2024impact} shows that if the node ordering is very different to the topological ordering of the true graph, it will likely adversely affect the accuracy of the learned graph. To counter this, lines 12-19 of Algorithm~\ref{algo:get-stable-order} attempt to improve the objective score, which typically also improves structural accuracy, by considering both $dec\_score\_order$ and its reverse. It chooses between these two orders by using HC to learn a graph with each order in lines 13 and 14, to test which results in the higher-scoring DAG. The topological order of the higher-scoring learned graph is returned as $stable\_order$. The results in Subsection~\ref{sub:ord_cat_bic} show that this empirical approach stabilises hill-climbing search \textbf{\textit{and}} improves the objective score of the learned graph. It also increases the structural accuracy when learning from categorical variables.

\section{Evaluation}
\label{sec:evaluation}

\renewcommand{\arraystretch}{1.2}
\setlength{\tabcolsep}{4pt} % Default is 6pt, reduce as needed
\begin{table}[h!t]
\centering
\rowcolors{2}{palegrey}{white}
\begin{footnotesize}
\begin{tabular}{l p{0.30\textwidth} lllllll}
\toprule
\thead{Network} & \thead[l]{Subject area} & \thead[l]{No. \\ of \\ nodes} & \thead[l]{No. \\ of \\arcs} & \thead[l]{Mean \& \\ (max.) \\ in-degree} & 
\thead[l]{Mean \& \\ (max.) \\ degree} & 
\thead[l]{Free \\ param- \\ -eters} \\
\midrule
Asia       & Simple patient diagnosis   &   8 &   8 & 1.00 (2) & 2.00 (4)   &      18 \\
Sports     & Football match outcomes    &   9 &  15 & 1.67 (2) & 3.33 (7)   &   1,049 \\
Sachs      & Protein signalling network &  11 &  17 & 1.55 (3) & 3.09 (7)   &     178 \\
Covid      & COVID in the UK            &  17 &  37 & 2.18 (5) & 4.35 (10)  &   7,834 \\
Child      & Lack of oxygen in babies   &  20 &  25 & 1.25 (2) & 2.50 (8)   &     230 \\
Insurance  & Car insurance risk         &  27 &  52 & 1.93 (3) & 3.85 (9)   &   1,008 \\
Property   & UK property investment     &  27 &  31 & 1.15 (3) & 2.30 (6)   &   3,056 \\
Diarrhoea  & Childhood diarrhoea        &  28 &  68 & 2.43 (8) & 4.86 (17)  &   1,716 \\
Water      & Waste water treatment      &  32 &  66 & 2.06 (5) & 4.12 (8)   &  10,083 \\
Mildew     & Mildew disease in wheat    &  35 &  46 & 1.31 (3) & 2.63 (5)   & 540,150 \\
Alarm      & Patient monitoring system  &  37 &  46 & 1.24 (4) & 2.49 (6)   &     509 \\
Barley     & Weed control in barley     &  48 &  84 & 1.75 (4) & 3.50 (8)   & 114,005 \\
Hailfinder & Forecasting severe weather &  56 &  66 & 1.18 (4) & 2.36 (17)  &   2,656 \\
Hepar2     & Diagnosing liver disorders &  70 & 123 & 1.76 (6) & 3.51 (19)  &   1,453 \\
Win95pts   & Computer diagnostics       &  76 & 112 & 1.47 (7) & 2.95 (10)  &     574 \\
Formed     & Prisoner reoffending risk  &  88 & 138 & 1.57 (6) & 3.14 (11)  &     912 \\
Pathfinder & Lymph-node diagnosis       & 109 & 195 & 1.79 (5) & 3.58 (106) &  71,890 \\
\bottomrule
\end{tabular}
\end{footnotesize}
\caption{Categorical variable networks used for evaluation.}
\label{tab:cat_networks}
\end{table}
\renewcommand{\arraystretch}{1.0}
\setlength{\tabcolsep}{6pt} % Default is 6pt, reduce as needed

The stability and accuracy of our approach is assessed using synthetic datasets generated from the seventeen categorical variable networks shown in Table~\ref{tab:cat_networks} and the seven continuous variable networks shown in Table~\ref{tab:con_networks}. These networks have between 8 and 109 nodes. The Building network is obtained from the BnRep repository \citep{leonelli2024bnrep}, Sports, Covid, Diarrhoea, Property and Formed are obtained from the Bayesys repository \citep{bayesysrepository} and the remainder from the bnlearn repository \citep{bnrepository}. These networks are commonly used in the literature to assess algorithms, and are largely expert-specified, generally representing causal networks found in the real world. Sample sizes of $10^2$, $10^3$, $10^4$ and $10^5$ are used to cover a typical range of sample sizes encountered in practical structure learning, including low-dimensional settings. The BIC score defined in equation~\ref{eqn:bic} is used throughout as the objective function for score-based algorithms and score-based phases of hybrid algorithms.

The learned graphs are evaluated using both their BIC score and their structural metrics. The BIC score scales with sample size, so we report the normalised BIC score which is the BIC score divided by sample size. The structural metrics compare the CPDAG of the true graph and with that of the learned graph since only observational data is being used. F1 and the Balanced Scoring Function (BSF) metric proposed in \cite{constantinou2019evaluating} are used as structural metrics. Both of these metrics have the advantage of not scaling with the network size, unlike SHD \citep{tsamardinos2006max}, which facilitates comparing performance across different networks.

\renewcommand{\arraystretch}{1.2}
\setlength{\tabcolsep}{4pt} % Default is 6pt, reduce as needed
\begin{table}[h!t]
\centering
\rowcolors{2}{palegrey}{white}
\begin{footnotesize}
\begin{tabular}{l p{0.32\textwidth} lllllll}
\toprule
\thead{Network} & \thead[l]{Subject area} & \thead[l]{No. \\ of \\ nodes} & \thead[l]{No. \\ of \\arcs} & \thead[l]{Mean \& \\ (max.) \\ in-degree} & 
\thead[l]{Mean \& \\ (max.) \\ degree} & 
\thead[l]{Free \\ param- \\ -eters} \\
\midrule
Sachs      & Protein signalling network  &  11 &  17 & 1.55 (3)  & 3.09 (7)  &  39 \\
Covid      & COVID in the UK             &  17 &  37 & 2.18 (5)  & 4.35 (10) &  71 \\
Building   & Damage to building concrete &  24 &  32 & 1.33 (4)  & 2.67 (5)  &  80 \\
Magic-NIAB & Genetic modelling in wheat  &  44 &  66 & 1.50 (9)  & 3.00 (10) & 154 \\
Ecoli70    & Gene activity in E. Coli    &  46 &  70 & 1.52 (4)  & 3.04 (11) & 162 \\
Magic-IRRI & Genetic modelling in wheat  &  64 & 102 & 1.59 (11) & 3.19 (11) & 230 \\
Arth150    & Gene activity in a plant    & 107 & 150 & 1.40 (6)  & 2.80 (20) & 364 \\
\bottomrule
\end{tabular}
\end{footnotesize}
\caption{Continuous variable networks used for evaluation.}
\label{tab:con_networks}
\end{table}
\renewcommand{\arraystretch}{1.0}
\setlength{\tabcolsep}{6pt} % Default is 6pt, reduce as needed

\noindent F1 ranges between 0.0 to 1.0 and is defined as
\begin{equation}
F1 = \frac{2 \times Precision \times Recall}{Precision + Recall}
\end{equation}
where Precision and Recall are defined as
\begin{equation}
Precision = \frac{TP}{TP + FP}, \, \, Recall = \frac{TP}{TP + FN}
\end{equation}
and TP is the number of True Positives, FP is False Positives and FN is False Negatives. We follow the approach adopted in the bnlearn package \citep{bnlearn} to compute these counts as shown in Table~\ref{tab:metrics}.

\begin{table}[H]
\centering
\begin{tabular}{ccccc}
\hline
\thead{Learned \\ graph} & \thead{Data-generating \\ graph} & \thead{True Positive \\ (TP)} & 
\thead{False positive \\ (FP)} & \thead{False Negative \\ (FN)} \\
\hline
$\longrightarrow$ & $\longrightarrow$ & 1 & 0 & 0 \\
\textbf{\large{\textemdash}} & \textbf{\large{\textemdash}} & 1 & 0 & 0 \\
$\longrightarrow$ & no edge & 0 & 1 & 0 \\
\textbf{\large{\textemdash}} & no edge & 0 & 1 & 0 \\
no edge & $\longrightarrow$ & 0 & 0 & 1 \\
no edge & \textbf{\large{\textemdash}} & 0 & 0 & 1 \\
$\longrightarrow$ & $\longleftarrow$ & 0 & 1 & 1 \\
$\longrightarrow$ & \textbf{\large{\textemdash}} & 0 & 1 & 1 \\
\textbf{\large{\textemdash}} & $\longrightarrow$ &  0 & 1 & 1 \\
\hline
\end{tabular}
\caption{The contribution to the True Positive, False Positive, False Negative counts (and hence F1 and BSF) resulting from different combinations of edges in the learned and data-generating graph.}
\label{tab:metrics}
\end{table}

BSF ranges from -1.0 to 1.0, with a value of 0.0 assigned to both the empty graph and fully connected graph. It incorporates all confusion matrix counts, including True Negatives (TNs), thereby considering the relative difficulty of identifying both the presence and absence of edges The BSF is defined as:
\begin{equation}
    \label{ref:sur_bsf}
    BSF = 0.5 \times \left( \frac{TP}{|E|} + \frac{TN}{|M|} - \frac{FP}{|M|} -  \frac{FP}{|E|} \right)
\end{equation}
where $|E|$ and $|M|$ represent the number of edges present and the number of edges absent (with reference to the complete graph) in the true graph respectively, so that:
\begin{equation}
    \label{ref:sur_absent_edges}
    |M| = 0.5 \times n \times (n - 1) - |E|
\end{equation}

To assess the stability of algorithms, each combination of sample size and network is repeated 25 times with the variable order, variable names, and row order in the dataset all randomised. The normalised BIC score and structural metrics are computed for each of these 25 experiments, and the standard deviation (SD) of the value reported as an indicator of the sensitivity of the algorithm to these randomised dataset artifacts. Given the large number of individual experiments involved, a time limit of three hours is imposed for structure learning for both categorical and continuous networks.

\section{Results}

\subsection{Comparison of Different Orderings}
\label{sub:ord_cat_bic}

\setlength{\tabcolsep}{2.5pt} % Default is 6pt, reduce as needed
\renewcommand{\arraystretch}{1.2}
\begin{table}[ht]
    \rowcolors{2}{palegrey}{white}
    \centering
    \begin{footnotesize}
    \begin{tabular}{p{0.18\textwidth} R{0.12\textwidth} R{0.13\textwidth} R{0.13\textwidth} R{0.11\textwidth} R{0.12\textwidth} R{0.11\textwidth} }
        \toprule
         &  \multicolumn{4}{c}{Tabu} & \multicolumn{2}{c}{HC} \\
        \cmidrule(r){2-5} \cmidrule(l){6-7}
        \textcolor{white}{x} \newline Metric &  Standard (unstable) & Decreasing score order & Increasing score order & Tabu- \newline -stable &
        Standard (unstable) & HC-Stable  \\
        \midrule
        Precision      &   0.4937 &   0.4741  &     0.5329 &   \B{0.5529} &   \W{0.4281} &     0.5029 \\
        Recall         &   0.4169 &   0.3993  &     0.4487 &   \B{0.4670} &   \W{0.3659} &     0.4248 \\
        F1             &   0.4426 &   0.4236  &     0.4776 &   \B{0.4976} &   \W{0.3844} &     0.4511 \\
        F1 SD          &   0.0828 &   0.0035  & \B{0.0030} &       0.0035 &   \W{0.0930} &     0.0032 \\
        BSF            &   0.5229 &   0.5070  &     0.5458 &   \B{0.5556} &   \W{0.4918} &     0.5353 \\
        Normalised BIC & -25.4818 & -25.4844  &   -25.4849 & \B{-25.4630} & \W{-25.5031} &   -25.4684 \\
        Norm. BIC SD   &   0.0445 & \B{0.000} & \B{0.0000} &   \B{0.0000} &   \W{0.0482} & \B{0.0000} \\
        \bottomrule
    \end{tabular}
    \end{footnotesize}
\caption{Mean values of structural metrics, normalised BIC and stability metrics averaged across all networks and sample sizes for standard and stable variants of Tabu and HC for the categorical variable networks. Best values are shown in bold, and worst values in bold red text.}
\label{tab:ord_cat_bic}
\end{table}
\renewcommand{\arraystretch}{1.0}
\setlength{\tabcolsep}{6pt} % Default is 6pt, reduce as needed

This subsection investigates the effect of using a stable ordering on the BIC score and structural metrics of the learned graphs, including their sensitivity to dataset artifacts. Table~\ref{tab:ord_cat_bic} presents the results obtained for the categorical networks. It compares the standard, unstable Tabu algorithm with the approach using the three stable orderings discussed in Subsection~\ref{sub:stabilising-tabu}: a) using an increasing score order, or b) a decreasing score order, or c) incorporating Algorithm~\ref{algo:get-stable-order} which tries both increasing and decreasing score order and uses the best one. This last approach is our proposed new algorithm which we call \say{Tabu-Stable}. Table~\ref{tab:ord_cat_bic} also presents results for the standard, unstable HC algorithm, and \say{HC-Stable}. The latter similarly incorporates Algorithm~\ref{algo:get-stable-order} to determine a stable order, but only allows positive score changes in the final hill-climbing phase.

Table~\ref{tab:ord_cat_bic} shows mean values of Precision, Recall, F1, BSF and normalised BIC averaged across all networks, sample sizes and dataset randomisations. It also presents the SD of both F1 and normalised BIC across the dataset randomisations, averaged across networks and sample sizes. The mean SD value reflects each algorithm's sensitivity to dataset artifacts such as variable ordering.

The stable ordering approaches in both HC and Tabu reduce mean F1 SD by around 30 and 25 times respectively, indicating that using a stable order improves the structural stability of the learned graph considerably. In the following subsections, we demonstrate that the remaining instability arises from duplicate variables. Tabu with a decreasing score order worsens the mean F1 value by 0.0190 whereas using an increasing score order improves it by 0.0350 over Tabu using variable order. However, Tabu-Stable produces the largest improvement in F1 of 0.0550 over Tabu, or about 12.5\%, as well as offering the best Precision and Recall values. This suggests that choosing the better of the decreasing and increasing score orders improves overall accuracy with categorical networks. HC using variable order is more unstable than Tabu, but HC-Stable reduces this instability considerably and increases mean F1 by 0.0667, around 17\%, over standard HC.

These accuracy improvements for categorical networks are also apparent using the alternative BSF metric. HC-Stable improves BSF by around 9\% over HC, and Tabu-Stable by approximately 6\% over Tabu. Tabu-Stable retains most of the accuracy improvement over HC-Stable that Tabu has over HC, suggesting that the accuracy improvement due to Tabu and that using a stable node order are additive in Tabu-Stable. These results show the improved structural accuracy of using Algorithm~\ref{algo:get-stable-order} to determine a stable order for categorical networks.

\begin{figure}[H]
    \centering
    \includegraphics[width=1.0\linewidth]{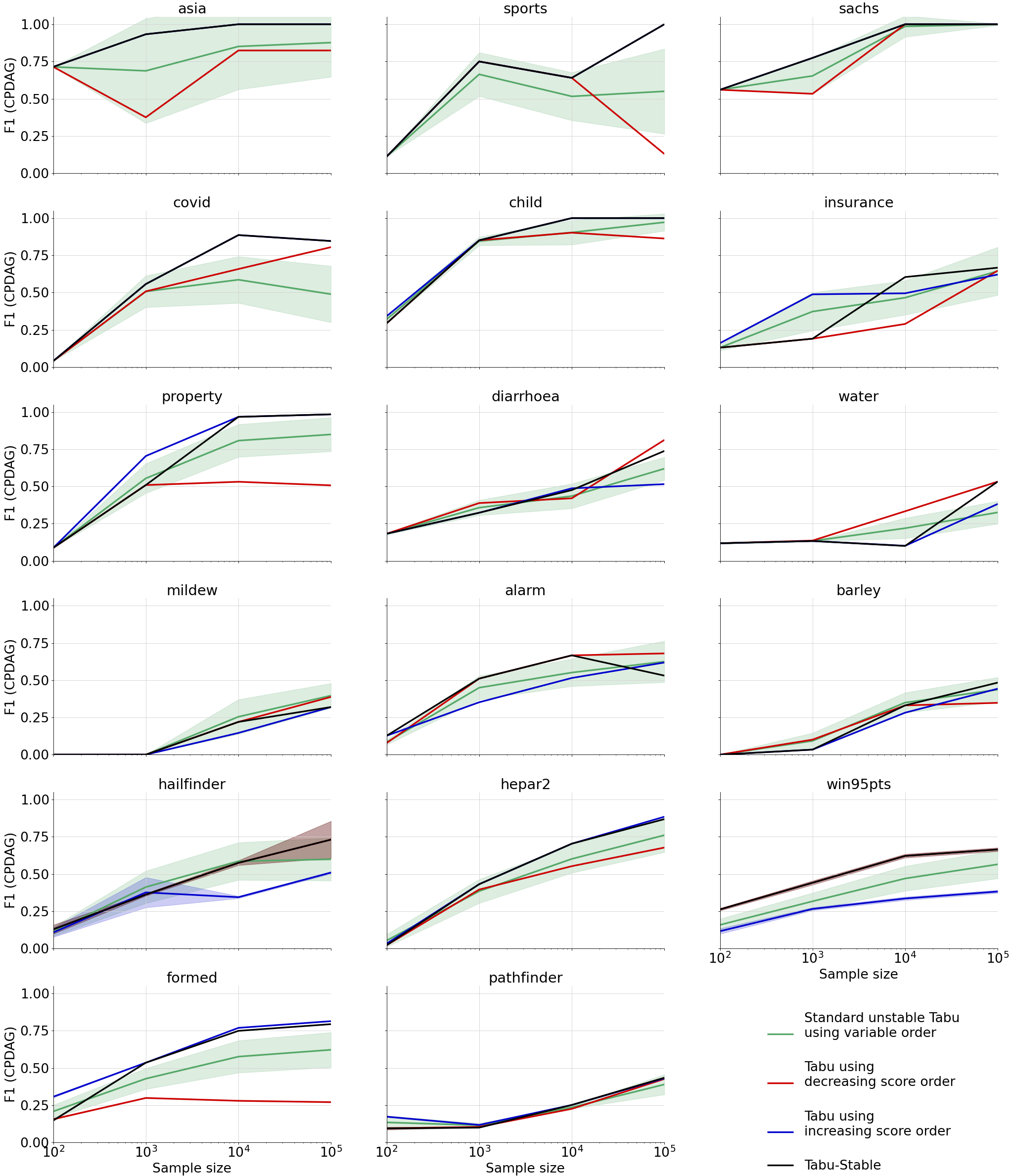}
    \caption{A comparison of the F1 CPDAG of the categorical variable graphs learned by Tabu using different node orderings: a) variable order, b) simple increasing or c) decreasing score order, and d) Tabu-Stable. 25 experiments with randomised variable names, variable order and row order are conducted for each of the four sample sizes for each network. Shading around lines indicates the SD of F1 values. Note that lines are drawn on the chart in the order shown in the key, so a particular line may be hidden where values are coincident.}
    \label{fig:order_f1}
\end{figure}

Figure~\ref{fig:order_f1} shows the F1 achieved by Tabu with the different orderings for each network at each sample size, with the shaded area around the lines indicating the SD of F1 values at each sample size. This instability is most pronounced for standard Tabu, shown in green, being considerable for most networks. The instability is substantially reduced with all three score-based orderings, although some instability is visible for the Hailfinder network which will be discussed in Subsection~\ref{sub:residual_instability}. 

The penultimate row in Table~\ref{tab:ord_cat_bic} shows the normalised BIC score averaged across all networks and sample sizes. It shows that using a decreasing or increasing score order in all cases produces a lower score than the standard, unstable algorithm, but Tabu-Stable and HC-Stable which use Algorithm~\ref{algo:get-stable-order} both improve the BIC score. Table~\ref{tab:score_cat_bic} provides a breakdown of the BIC scores obtained for each network. In addition to Tabu, Tabu-Stable, HC, and HC-Stable, it shows the BIC scores for two baseline graphs: the empty graph, and the true graph that was used to generate the synthetic dataset. Tabu-Stable achieves the highest BIC score for 16/17 of the networks, but a lower score than standard Tabu for the Barley network. The empty graph produces the worst (lowest) score in 8/17 networks which is unsurprising since it is equivalent to assuming that all the network variables are independent. We also observe that the \textit{\textbf{true graph}} receives the lowest score in 9/17 networks, especially in highly parameterised ones. This likely results from a poor fit to data at low sample sizes, combined with a high complexity penalty. Under these conditions, score-based approaches are unlikely to accurately recover the true graph.

\setlength{\tabcolsep}{3pt} % Default is 6pt, reduce as needed
\renewcommand{\arraystretch}{1.2}
\begin{table}[!ht]
    \rowcolors{2}{palegrey}{white}
    \centering
    \begin{footnotesize}
    \begin{tabular}{p{0.15\textwidth} R{0.12\textwidth} R{0.13\textwidth} R{0.13\textwidth} R{0.12\textwidth} R{0.11\textwidth} R{0.13\textwidth} }
        \toprule
         & \multicolumn{2}{c}{Tabu} & \multicolumn{2}{c}{HC} & \multicolumn{2}{c}{Baselines} \\
        \cmidrule(r){2-3} \cmidrule(lr){4-5} \cmidrule(l){6-7}
        \rowcolor{white} \textcolor{white}{x} \newline Metric &  Standard (unstable) & Tabu-Stable & Standard (unstable) & HC-stable & Empty graph & True graph  \\
        \midrule
        Asia       &     -2.3383  &   \B{-2.3374} &   -2.3437 &     -2.3383  &  \W{-3.0625} &      -2.3537  \\
        Sports     &    -11.7874  &  \B{-11.7538} &  -11.8079 &    -11.7549  &    -13.5246  &  \W{-16.9389} \\
        Sachs      &     -7.6888  &   \B{-7.6862} &   -7.6968 &     -7.6901  &  \W{-9.3592} &      -8.2304  \\
        Covid      &    -14.3985  &  \B{-14.3030} &  -14.4524 &    -14.3081  &    -20.7702  &  \W{-63.1447} \\
        Child      &    -13.0724  &  \B{-13.0642} &  -13.1136 &    -13.0811  & \W{-17.2956} &     -13.3947  \\
        Insurance  &    -14.6687  &  \B{-14.6546} &  -14.6990 &    -14.8882  & \W{-21.5860} &     -17.8317  \\
        Property   &    -27.6299  &  \B{-27.6288} &  -27.6374 & \B{-27.6288} &    -40.3758  &  \W{-43.2863} \\
        Diarrhoea  &    -19.9769  &  \B{-19.9689} &  -19.9812 &    -19.9696  &    -21.9777  &  \W{-29.9880} \\
        Water      &    -13.6683  &  \B{-13.6631} &  -13.6700 &    -13.6636  &    -17.0671  &  \W{-22.8180} \\
        Mildew     &    -54.4729  &  \B{-54.4527} &  -54.4897 &    -54.4579  &    -63.5583  & \W{-674.7136} \\
        Alarm      &    -12.1061  &  \B{-12.0909} &  -12.1311 &    -12.0977  & \W{-20.8005} &     -13.2097  \\
        Barley     & \B{-65.8108} &     -65.9052  &  -65.8643 &    -65.9064  &    -83.3012  & \W{-562.0477} \\
        Hailfinder &    -54.0439  &  \B{-54.0042} &  -54.0661 &    -54.0150  & \W{-70.5261} &     -65.0452  \\
        Hepar2     &    -33.7398  &  \B{-33.7376} &  -33.7464 &    -33.7419  &    -35.9661  &  \W{-41.8153} \\
        Win95pts   &    -10.1453  &  \B{-10.0883} &  -10.1489 &    -10.0899  & \W{-19.1424} &     -11.8806  \\
        Formed     &    -43.9545  &  \B{-43.8817} &  -43.9722 &    -43.8835  & \W{-60.5757} &     -46.9654  \\
        Pathfinder &    -33.6885  &  \B{-33.6506} &  -33.7318 &    -33.6686 &     -63.3976  & \W{-254.3176} \\
        \bottomrule
    \end{tabular}
    \end{footnotesize}
\caption{Mean values of the normalised BIC score averaged across all sample sizes for each categorical network for standard and stable variants of Tabu and HC, as well as two baseline graphs. Best values are shown in bold, and worst values in bold red text.}
\label{tab:score_cat_bic}
\end{table}
\renewcommand{\arraystretch}{1.0}
\setlength{\tabcolsep}{6pt} % Default is 6pt, reduce as needed

The final row of Table~\ref{tab:ord_cat_bic} shows the SD of the normalised BIC score which we see is zero for the approaches using a stable order but non-zero for the standard HC and Tabu algorithms. Thus, the approaches discussed in Subsection~\ref{sub:stabilising-tabu} completely eliminate the instability in the BIC score. 

Table~\ref{tab:ord_con_bic} provides the analogous results to Table~\ref{tab:ord_cat_bic} but for the seven continuous variable networks. For these networks we again see that the stable order based approaches completely eliminate the instability in BIC score \textbf{\textit{and}} F1, and that Tabu-Stable again produces the best normalised BIC score averaged over all networks. However, the stable order approaches \textbf{\textit{worsen}} F1 for continuous variable networks, with Tabu-Stable achieving a mean F1 of 0.5284, compared to 0.5621 for standard Tabu, a decrease of around 6\%. The F1 for HC-Stable is slightly better than that for HC. The BSF metric offers more favourable results; Tabu-Stable decreases BSF by only around 1\% over Tabu, and that for HC-Stable is around 1.5\% higher than HC.

\setlength{\tabcolsep}{2.5pt} % Default is 6pt, reduce as needed
\renewcommand{\arraystretch}{1.2}
\begin{table}[ht]
    \rowcolors{2}{palegrey}{white}
    \centering
    \begin{footnotesize}
    \begin{tabular}{p{0.18\textwidth} R{0.12\textwidth} R{0.13\textwidth} R{0.13\textwidth} R{0.11\textwidth} R{0.12\textwidth} R{0.11\textwidth} }
        \toprule
         &  \multicolumn{4}{c}{Tabu} & \multicolumn{2}{c}{HC} \\
        \cmidrule(r){2-5} \cmidrule(l){6-7}
        \textcolor{white}{x} \newline Metric &  Standard (unstable) & Decreasing score order & Increasing score order & Tabu- \newline -stable &
        Standard (unstable) & HC-Stable  \\
        \midrule
        Precision      &     0.4937 & \B{0.5194} & \W{0.4530} &       0.5043 &       0.5033 &     0.5098 \\
        Recall         & \W{0.4169} & \B{0.5955} &     0.5137 &       0.5661 &       0.5740 &     0.5699 \\
        F1             & \B{0.5621} &     0.5498 & \W{0.4760} &       0.5284 &       0.5303 &     0.5324 \\
        F1 SD          &     0.0998 & \B{0.0000} & \B{0.0000} &   \B{0.0000} &   \W{0.1002} & \B{0.0000} \\
        BSF            &     0.7533 & \B{0.7598} & \W{0.7220} &       0.7474 &       0.7295 &     0.7419 \\
        Normalised BIC &   -52.3305 &   -52.3299 &   -52.3338 & \B{-52.3230} & \W{-52.3368} &   -52.3268 \\
        Norm. BIC SD   &     0.0151 & \B{0.0000} & \B{0.0000} &   \B{0.0000} &   \W{0.0160} & \B{0.0000} \\
        \bottomrule
    \end{tabular}
    \end{footnotesize}
\caption{Mean values of structural and inference metrics averaged across all networks and sample sizes for standard and stable variants of Tabu and HC for the continuous networks. Best values are shown in bold, and worst values in bold red text.}
\label{tab:ord_con_bic}
\end{table}
\renewcommand{\arraystretch}{1.0}
\setlength{\tabcolsep}{6pt} % Default is 6pt, reduce as needed

Table~\ref{tab:score_con_bic} provides the BIC scores for each network for the standard and stable versions of Tabu and HC, as well as the empty and true graph. Tabu-Stable produces the highest-scoring graph in 6/7 networks, and is always higher scoring than the true graph. The empty graph always has the worst score for these continuous networks. The BIC score achieves its maximum value for the true graph only in the asymptotic limit as the sample size approaches infinity. Here, the learned graphs have higher BIC scores than the true graph, which may explain why the stabilised algorithms, while improving BIC scores, result in a decrease in F1 performance.

\setlength{\tabcolsep}{2.5pt} % Default is 6pt, reduce as needed
\renewcommand{\arraystretch}{1.2}
\begin{table}[H]
    \rowcolors{2}{palegrey}{white}
    \centering
    \begin{footnotesize}
    \begin{tabular}{p{0.15\textwidth} R{0.12\textwidth} R{0.13\textwidth} R{0.13\textwidth} R{0.12\textwidth} R{0.13\textwidth} R{0.12\textwidth} }
        \toprule
         & \multicolumn{2}{c}{Tabu} & \multicolumn{2}{c}{HC} & \multicolumn{2}{c}{Baselines} \\
        \cmidrule(r){2-3} \cmidrule(lr){4-5} \cmidrule(l){6-7}
        \rowcolor{white} \textcolor{white}{x} \newline Metric &  Standard (unstable) & Tabu-Stable & Standard (unstable) & HC-stable & Empty graph & True graph  \\
        \midrule
        Sachs      &    -67.9587  & \B{-67.9577} &  -67.9609 &    -67.9591  & \W{ -73.1885} &  -67.9874 \\
        Covid      &    -90.0674  & \B{-90.0655} &  -90.0705 &    -90.0701  & \W{-101.3074} &  -90.1249 \\
        Building   &     -2.3933  & \B{ -2.3826} &   -2.4021 &     -2.3838  & \W{ -40.2316} &   -2.3891 \\
        Magic-NIAB &    -48.9989  & \B{-48.9892} &  -49.0019 &    -48.9916  & \W{ -50.4217} &  -49.2406 \\
        Magic-IRRI &    -76.7411  & \B{-76.7333} &  -76.7451 &    -76.7361  & \W{ -83.6172} &  -77.2543 \\
        Ecoli70    & \B{-42.2674} &    -42.2743  &  -42.2815 &    -42.2775  & \W{ -74.9003} &  -42.3818 \\
        Arth150    &    -37.8868  & \B{-37.8581} &  -37.8956 &    -37.8694  & \W{ -61.5820} &  -38.6545 \\
        
        \bottomrule
    \end{tabular}
    \end{footnotesize}
\caption{Mean values of the normalised BIC score averaged across all sample sizes for each continuous network for standard and stable variants of Tabu and HC, as well as two baseline graphs. Best values are shown in bold, and worst values in bold red text.}
\label{tab:score_con_bic}
\end{table}
\renewcommand{\arraystretch}{1.0}
\setlength{\tabcolsep}{6pt} % Default is 6pt, reduce as needed

\subsection{Analysis of the Residual Instability in Categorical Networks with Tabu-Stable}
\label{sub:residual_instability}

\begin{table}[ht]
\centering
\begin{tabular}{ccccccc}
\toprule
\thead{Sample size} & \thead{ \\ hailfinder} & \thead{ \\ win95pts} & \thead{ \\ formed } & \thead{ \\ pathfinder} 
& \thead{ \\ hailfinder2} & \thead{ \\ win95pts2 }  \\
\midrule
$10^2$ & 0.0301 & 0.0081 & 0.0041 & 0.0093 & 0.0000 & 0.0077 \\
$10^3$ & 0.0081 & 0.0135 & 0.0000 & 0.0000 & 0.0000 & 0.0000 \\
$10^4$ & 0.0157 & 0.0123 & 0.0000 & 0.0000 & 0.0000 & 0.0000 \\
$10^5$ & 0.1245 & 0.0120 & 0.0000 & 0.0000 & 0.0000 & 0.0000 \\
\bottomrule
\end{tabular}
\caption{F1 SD over 25 random orderings at different sample sizes using Tabu-Stable for the four networks with residual sensitivity to variable order, and for modified versions of Hailfinder and Win95pts where identical (or duplicate) variables are prevented.}
\label{tab:remaining}
\end{table}

Tabu-Stable returns a F1 SD of 0.000 at all sample sizes for all 7 continuous networks and for 13 out of the 17 categorical networks. Table~\ref{tab:remaining} provides a breakdown of the F1 SD by sample size using Tabu-Stable for the four categorical networks where some structural instability remains. Formed and Pathfinder only have residual instability at a sample size of 100. However, Hailfinder and Win95pts retain some instability at all sample sizes. This is because both networks have some local structures and CPT values that deterministically create identical values for pairs of variables at all sample sizes. The last two columns in Table~\ref{tab:remaining} show results for versions of these networks that are modified slightly to remove these deterministic relationships. Hailfinder2 and Win95pts2 have had one node and two arcs removed respectively to achieve this. Table~\ref{tab:remaining} shows that instability has been removed completely for Hailfinder2 and only remains at the smallest sample size for Win95pts2 because some variables are identical.

\subsection{Comparing Tabu-Stable and HC-Stable with other algorithms}
\label{sub:other_algorithms}

\subsubsection{Categorical Networks}
\label{subsub:other_algorithms_categorical}

\begin{figure}[htp]
    \centering
    \includegraphics[width=0.97\linewidth]{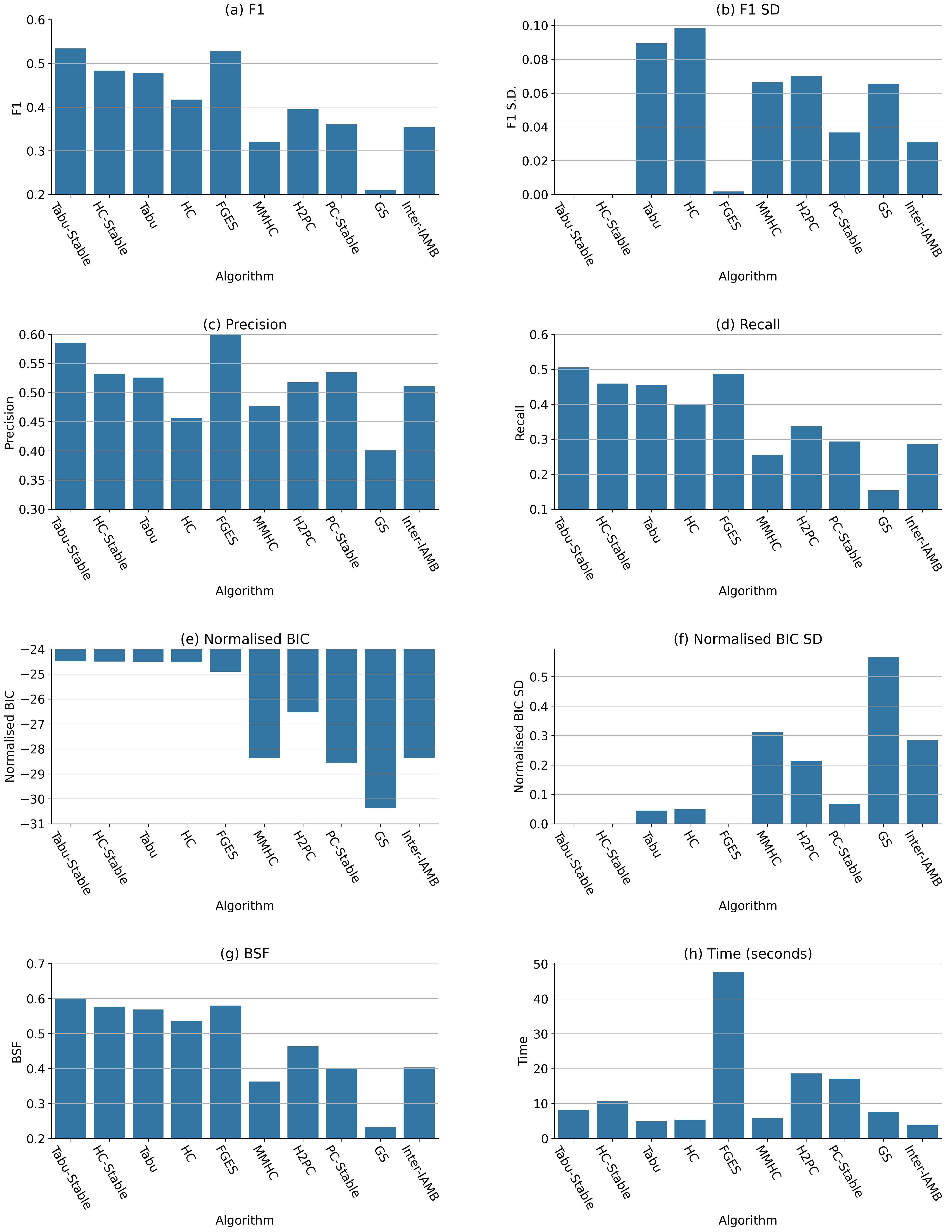}
    \caption{Mean values of structural, inference and stability metrics across categorical networks and sample sizes for different algorithms.}
    \label{fig:algos_cat_bic}
\end{figure}

Figure~\ref{fig:algos_cat_bic} compares the mean structural, normalised BIC score and stability metrics for the categorical networks achieved by different algorithms with those obtained by HC-Stable and Tabu-Stable. The implementation of FGES from the Tetrad package \citep{ramsey2018tetrad}, and MMHC, H2PC, PC-Stable, GS and Inter-IAMB from the bnlearn package \citep{bnlearn} are used. Results for standard Tabu and HC using variable order are also included. The results use the variants of Hailfinder and Win95pts discussed in Subsection~\ref{sub:residual_instability} which avoid identical variables.

Additionally, the bnlearn algorithms reject datasets which contain any variable that has the same value for all rows, so the cases where this occurs are also excluded from this comparison. These are the datasets with 100 rows for the Insurance, Water, Barley, Hailfinder, Win95pts, Formed and Pathfinder networks. Excluding experiments where datasets contain identical or single-valued variables means that the algorithms have potential to achieve full stability, provided they are \textbf{\textit{truly}} stable. Thus, all algorithms are compared across the same experiments, with the exception that FGES failed to complete within 3 hours for sample sizes of $10^4$ and $10^5$ for Hailfinder and Pathfinder, that is, for 4/58 experiments. We ignore these failure cases when computing the results shown in Figure~\ref{fig:algos_cat_bic}. This introduces some bias into the results but, as we show below, when we impute missing results instead, this has a relatively small effect and did not change the ranking of algorithms. Moreover, result imputation introduces its own biases; for example, making FGES appear more unstable than it actually is.

Tabu-Stable and HC-Stable are the only algorithms to produce a mean F1 SD of zero, by completely eliminating structural instability. FGES is found to be \textbf{\textit{almost}} structurally stable with a mean F1 SD of 0.0018. The other algorithms, including PC-Stable which has a mean F1 SD of 0.0367, all exhibit considerable structural instability. Moreover, Tabu-Stable also offers the highest mean F1 of 0.5341 - this value is higher than that quoted in Table~\ref{tab:ord_cat_bic} because the datasets not considered in this set of experiments, due to duplicate or single-valued variables, tend to be those with lowest sample sizes that also tend to produce lower F1 values. FGES produces the second highest mean F1 of 0.5278 when it's failure cases are ignored, or 0.5188 if its failure cases are assigned the mean F1 value obtained across all the other algorithms. FGES provides the best Precision, 0.0194 better than Tabu-Stable, but the latter provides the best Recall, 0.0182 better than FGES. Tabu-Stable also achieved the highest mean BSF of 0.5995, with FGES second at 0.5799, closely followed by HC-Stable at 0.5772.

Tabu-Stable achieved the best mean normalised BIC score of -24.4970, with HC-Stable close behind at -24.5032, and FGES third at -24.9100. The hybrid and constraint-based algorithms obtained considerably worse mean BIC scores in the range -26.5373 to -30.3693, which is expected since they are not orientated to fully maximise an objective score. As noted previously, Tabu-Stable and HC-Stable achieved zero BIC SD, that is, complete score stability too. FGES had very stable BIC scores with a normalised BIC SD of just 0.0006. Standard Tabu, HC and PC-Stable had normalised BIC SDs of 0.0452, 0.0494 and 0.0679 respectively. GS, Inter-IAMB, MMHC and H2PC demonstrated much larger BIC instability with normalised BIC SDs between 0.2144 and 0.5659. Tabu-Stable was around 68\% slower than Tabu, and HC-Stable nearly twice as slow as HC, but were generally competitive with the runtimes of the other algorithms; for example, Tabu-Stable was around twice as fast as PC-Stable and H2PC, and much faster than FGES. FGES was the slowest algorithm tested with a mean run time around 50 seconds, but most of this larger mean execution time was due to extended runtimes on a few networks, notably Mildew and Hailfinder2. Tabu-Stable is faster than HC-Stable, and Tabu is faster than HC, which is surprising since the Tabu approach generally performs more iterations than plain HC. Tabu may be faster because its \textit{tabulist} prevents it from considering some recently visited high-scoring graphs and thereby avoids relatively costly acyclicity checks for those graphs.

\subsubsection{Continuous Networks}
\label{subsub:other_algorithms_continuous}

\begin{figure}[htp]
    \centering
    \includegraphics[width=0.97\linewidth]{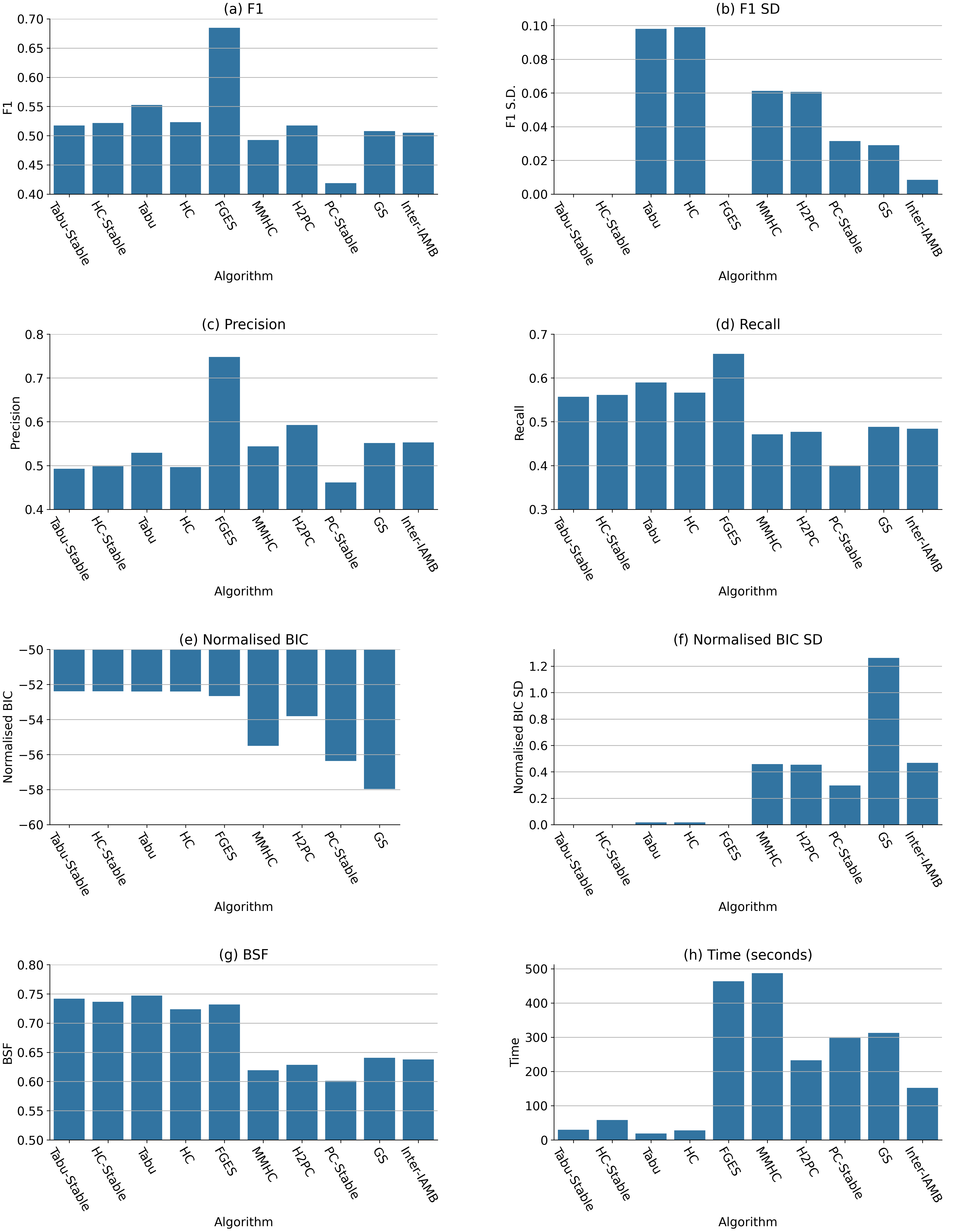}
    \caption{Mean values of structural, inference and stability metrics across \textbf{continuous} networks and sample sizes for different algorithms.}
    \label{fig:algos_con_bic}
\end{figure}

Figure~\ref{fig:algos_con_bic} presents equivalent algorithm comparisons for the seven continuous networks. Once again, we show averages across all sample sizes and networks, but ignore individual failure cases for a particular algorithm, sample size and network; these failure cases are discussed further below. Tabu-Stable, HC-Stable and FGES all have F1 and BIC SD values of 0.0 demonstrating complete structural and score stability with these continuous networks. The F1 SD for Inter-IAMB is modest at 0.0084, but the other algorithms have F1 SD values between 0.0290 and 0.0990 indicating considerable structural instability. On the other hand, standard Tabu and HC have relatively low values for normalised BIC SD of 0.0168 and 0.0178 respectively, with the other algorithms exhibiting large BIC instability with SD ranging from 0.2963 to 1.2635. It is perhaps to be expected that the score-based algorithms have the more stable BIC scores since they aim to maximise the objective score.

Tabu-Stable achieves the best overall mean BIC score of -52.3854, with HC-Stable, Tabu and HC next best at -52.3898, -52.3943 and -52.4009 respectively. FGES has the fifth best score of -52.6603, with the hybrid and constraint-based algorithms having considerably worse scores. This finding remains the same if we instead impute BIC values for failed experiments, though the ranking within the constraint and hybrid algorithms according to BIC score alters.

The networks and sample sizes where experiments failed to complete within the time limit are shown in Table~\ref{tab:failure_cases}. In some cases, runtimes were far in excess of three hours for a particular sample size and network, so that none of the randomised variations produced a result within the time limit which is shown by a red cross in Table~\ref{tab:failure_cases}. In other cases, marked by a blue cross in the table, some of the randomised datasets produced a result within the time limit and others did not. Thus, the execution runtime can also be significantly affected by dataset artifacts such as variable ordering in the algorithms shown in Table~\ref{tab:failure_cases}.

\setlength{\tabcolsep}{4pt} % Default is 6pt, reduce as needed
\begin{table}[ht]
\centering
\begin{footnotesize}
\begin{tabular}{llcccccc}
\toprule
Network    &  Sample Size & FGES & MMHC & H2PC & PC-Stable & GS  & Inter-IAMB \\
\midrule
Magic-NIAB & 100,000      &      &      & \rx  &                 & \rx & \rx  \\
Magic-IRRI & 100,000      & \bx  &      & \rx  &                 & \rx & \rx  \\
Ecoli70    & 100,000      & \bx  &      &      &                 &     &      \\
Arth150    &  10,000      & \bx  &      & \rx  &                 & \bx &      \\
Arth150    & 100,000      & \rx  & \rx  & \rx  & \rx             & \rx & \rx  \\
\bottomrule
\end{tabular}
\end{footnotesize}
\caption{Networks and sample sizes at which algorithms failed to learn a graph within the 3 hour time limit. Combinations where all dataset randomisations failed to learn a graph are marked with a red cross, and those where only some of the randomised datasets failed are marked with a blue cross.}
\label{tab:failure_cases}
\end{table}
\setlength{\tabcolsep}{6pt} % Default is 6pt, reduce as needed

Another complication when comparing the algorithms is that constraint-based algorithms often produce a PDAG which cannot be extended into a DAG, and therefore cannot be represented as a CPDAG nor assigned an objective score. Table~\ref{tab:nonex_cases} shows the percentage of runs that completed within the time limit but which produced a PDAG that could not be extended. Note that, in general, this failure to learn extendable graphs occurred across most networks and sample sizes. Typically, some of the learned graphs for a particular network and sample size are extendable and others not, so this represents yet another feature of the learned graph that is sensitive to dataset artifacts. Inter-IAMB is particularly prone to producing non-extendable PDAGs and did not produce \textit{\textbf{any}} extendable PDAGs for the Covid, Magic-NIAB, Magic-IRRI and Arth150 continuous networks. This means that the mean BIC score for Inter-IAMB for continuous networks is not comparable with the other algorithms, and hence Inter-IAMB is not shown in Figure~\ref{fig:algos_cat_bic}(e).     

\begin{table}[ht]
\centering
\begin{small}
\begin{tabular}{lcccccc}
\toprule
Network     & PC-Stable   & GS   & Inter-IAMB \\
\midrule
Categorical & 12.9 &  2.4 & 16.4       \\
Continuous  & 28.3 & 67.6 & 68.0       \\
\bottomrule
\end{tabular}
\end{small}
\caption{Percentage of learned graphs that were non-extendable, and therefore could not be allocated an objective score.}
\label{tab:nonex_cases}
\end{table}

Unlike the results for the categorical networks, the better objective scores obtained by Tabu-Stable \textbf{\textit{are not reflected}} in better structural accuracy. The mean F1 value for Tabu-Stable is 0.5175, 0.0350 \textbf{\textit{worse}} than Tabu, and HC-Stable at 0.5217 is 0.0014 worse than HC. FGES produces the best mean F1 value at 0.6849, with Tabu and HC next, followed by HC-Stable and Tabu-Stable. H2PC has the same F1 value as Tabu-Stable, with the other hybrid and constraint-based algorithms close behind, except for PC-Stable which is considerably worse. If we impute F1 values for failed experiments then the mean F1 values for H2PC, GS and Inter-IAMB rise by around 0.02 to 0.03 raising them above HC-Stable and Tabu-Stable but not altering the picture dramatically.

Tabu-Stable and HC-Stable fared better according to the BSF structural metric, where Tabu-Stable has the second highest BSF value of 0.7420, slightly ahead of FGES at 0.7319, but worse than standard Tabu at 0.7473. The hybrid and constraint-based algorithms are substantially less accurate than the score-based algorithms according to the BSF metric.

\setlength{\tabcolsep}{6pt} % Default is 6pt, reduce as needed
\renewcommand{\arraystretch}{1.2}
\begin{table}[H]
    \rowcolors{2}{palegrey}{white}
    %% \centering
    \begin{footnotesize}
    \begin{tabular}{p{0.15\textwidth} p{0.15\textwidth} p{0.19\textwidth} p{0.15\textwidth} p{0.19\textwidth} }
        \toprule
         & \multicolumn{2}{c}{Categorical networks} & \multicolumn{2}{c}{Continuous networks} \\
        \cmidrule(r){2-3} \cmidrule(lr){4-5} 
        Algorithm   &  Mean of \newline node degree & Mean of \newline node degree SD & 
        Mean of \newline node degree & Mean of \newline node degree SD \\
        \midrule
        True graph  & 3.2328   & n/a        & 3.0288  & n/a        \\
        HC          & 2.5603   & 0.1041     & 3.7771  & 0.2478     \\
        HC-Stable   & 2.4917   & 0.0000     & 3.6392  & 0.0000     \\
        Tabu        & 2.5155   & 0.0882     & 3.6871  & 0.2335     \\
        Tabu-Stable & 2.4618   & 0.0000     & 3.6334  & 0.0000     \\
        FGES        & 2.2547   & 0.0008     & 2.6061  & 0.0000     \\
        MMHC        & 1.4307   & 0.0659     & 2.6017  & 0.0242     \\
        H2PC        & 1.7904   & 0.0512     & 2.3539  & 0.0020     \\
        PC-Stable   & 1.4690   & 0.0000     & 2.5813  & 0.0008     \\
        GS          & 1.0265   & 0.1301     & 2.7062  & 0.0600     \\
        Inter-IAMB  & 1.5578   & 0.0375     & 2.7082  & 0.0095     \\
        \bottomrule
    \end{tabular}
    \end{footnotesize}
\caption{Mean of node degree and mean of node degree SD for categorical and continuous networks for the true graph and graphs learned by the different algorithms.}
\label{tab:densities}
\end{table}
\renewcommand{\arraystretch}{1.0}
\setlength{\tabcolsep}{6pt} % Default is 6pt, reduce as needed

The BIC objective score seems to be a better indicator of correct structure when used with categorical data than it is with continuous data, a finding also reported in \cite{constantinou2023open} and \cite{scutari2019learns}. Table~\ref{tab:densities} suggests a reason why this might be so. It shows the mean of node degree and mean of node degree SD of the learned graph for each algorithm for the categorical and continuous networks, where node degree is the sum of in-degree and out-degree for a node. The node degree reflects the density of the graph skeleton, and the node degree SD the instability of the skeleton. The mean node degree of the true graphs is also shown: 3.2328 for the categorical networks and 3.0288 for the continuous networks. With the categorical networks, all algorithms tend to learn graphs with a \textit{\textbf{lower}} node degree, that is, density, than the true graphs, but the HC/Tabu family of algorithms learn graphs with a density closest to the true value, contributing to their good structural metrics. Conversely, all the algorithms learn graphs with a \textit{\textbf{higher}} mean density than the true mean density for the continuous networks. Once again, the HC/Tabu family of algorithms learn the densest graphs, but this now means they are furthest away from the true density, contributing to poorer structural metrics.

Table~\ref{tab:densities} also shows that GS, Inter-IAMB, MMHC and H2PC may have relatively high values for the density SD, indicating that data artifacts can also affect the skeleton of the learned graph. Interestingly, PC-Stable has low values for the density SD suggesting that the implementation we are using may have a stable skeleton determination phase, and the structural instability seen in the F1 SD value stems from arc orientation instability in the v-structure or orientation propagation phases of PC-Stable.

\section{Concluding Remarks}
\label{sec:concluding_remarks}

We present HC-Stable and Tabu-Stable, two new hill-climbing algorithms that eliminate instability in learning categorical and continuous variable networks, provided datasets do not contain variables with identical sequences of values. Unlike many well-known algorithms, which are sensitive to dataset artifacts such as variable ordering, our approach ensures that the objective score, structure, and runtime of the learned graph are unaffected by dataset artifacts such as variable ordering.

We show that this instability in standard algorithms can affect the learned graph's objective score, skeleton, arc orientations, and even runtime. For some constraint-based algorithms, it may also prevent converting the learned PDAG into a usable CPDAG or DAG. Our method overcomes these issues by first determining a stable node order based on objective scores, which is then used in subsequent hill-climbing to resolve arc orientation when optimal objective scores are tied.

These new algorithms demonstrate clear benefits:

\begin{itemize}
\item \textbf{Stability}: HC-Stable and Tabu-Stable produce the same results regardless of variable ordering, naming, or row ordering for a given algorithm, network and sample size.
\item \textbf{Performance}: Tabu-Stable achieves the highest normalised BIC scores across all networks evaluated, and both algorithms improve the accuracy of learned graphs for categorical networks.
\item \textbf{Efficiency}: Despite additional computations, their runtimes remain competitive, significantly outperforming FGES, the only other very stable algorithm here, for both categorical and continuous networks.
\item \textbf{Scalability}: FGES, the constraint-based and hybrid algorithms all produce cases where runtime is unusually high for particular networks whereas the new algorithms proposed in this study demonstrate consistent runtimes across all networks investigated.
\end{itemize}

For networks with continuous variables, we observe that our proposed algorithms sometimes decrease structural accuracy even when BIC scores improve. This highlights a limitation of score-based algorithms: they are designed to optimise a specific objective score, but this does not always translate into more accurate graphs, particularly when sample sizes are modest. In this study, this issue is evident as Tabu-Stable frequently produces graphs with higher BIC scores than the true graph, even when the learned graph is quite inaccurate.

The methodology is flexible and can incorporate alternative methods for determining node order, such as that of \cite{behjati2020improved}, provided no dataset artifacts influence the ordering. We recommend Tabu-Stable for practical applications and algorithm comparisons, given its stability, simplicity, and robust performance. The algorithm, along with instructions for its use and for reproducing all results and charts presented here, are freely available on GitHub at \url{https://github.com/causal-iq/discovery}. Moreover, we emphasise that stability considerations should be integral to algorithm design and evaluation, as instability can bias benchmarks and practical outcomes. Investigating the causes of instability can further inspire the development of new and improved algorithms.

%% \appendix uncomment if appendices required

\bibliographystyle{plain}

\begin{thebibliography}{10}

\bibitem{bartlett2017integer}
Mark Bartlett and James Cussens.
\newblock Integer linear programming for the bayesian network structure learning problem.
\newblock {\em Artificial Intelligence}, 244:258--271, 2017.

\bibitem{behjati2020improved}
Shahab Behjati and Hamid Beigy.
\newblock Improved k2 algorithm for bayesian network structure learning.
\newblock {\em Engineering Applications of Artificial Intelligence}, 91:103617, 2020.

\bibitem{bouckaert1995bayesian}
Remco~Ronaldus Bouckaert.
\newblock {\em Bayesian belief networks: from construction to inference}.
\newblock PhD thesis, University of Utrecht, 1995.

\bibitem{broom2012model}
Bradley~M Broom, Kim-Anh Do, and Devika Subramanian.
\newblock Model averaging strategies for structure learning in bayesian networks with limited data.
\newblock {\em BMC bioinformatics}, 13:1--18, 2012.

\bibitem{cai2017bayesian}
Baoping Cai, Lei Huang, and Min Xie.
\newblock Bayesian networks in fault diagnosis.
\newblock {\em IEEE Transactions on industrial informatics}, 13(5):2227--2240, 2017.

\bibitem{chickering2002optimal}
David~Maxwell Chickering.
\newblock Optimal structure identification with greedy search.
\newblock {\em Journal of machine learning research}, 3(Nov):507--554, 2002.

\bibitem{colombo2014order}
Diego Colombo and Marloes~H Maathuis.
\newblock Order-independent constraint-based causal structure learning.
\newblock {\em Journal of Machine Learning Research}, 15:3921--3962, 2014.

\bibitem{bayesysrepository}
A~C Constantinou, Y~Liu, K~Chobtham, Z~Guo, and N~K Kitson.
\newblock {The Bayesys data and Bayesian network repository v1.5}.
\newblock {http://bayesian-ai.eecs.qmul.ac.uk/bayesys/}, 2024.
\newblock Bayesian Artificial Intelligence research lab, Queen Mary University of London, London, UK.

\bibitem{constantinou2023open}
Anthony Constantinou, Neville~K Kitson, Yang Liu, Kiattikun Chobtham, Arian~Hashemzadeh Amirkhizi, Praharsh~A Nanavati, Rendani Mbuvha, and Bruno Petrungaro.
\newblock Open problems in causal structure learning: A case study of covid-19 in the uk.
\newblock {\em Expert Systems with Applications}, 234:121069, 2023.

\bibitem{constantinou2019evaluating}
Anthony~C Constantinou.
\newblock {Evaluating structure learning algorithms with a balanced scoring function}.
\newblock {\em arXiv preprint arXiv:1905.12666}, 2019.

\bibitem{cooper1992bayesian}
Gregory~F Cooper and Edward Herskovits.
\newblock A bayesian method for the induction of probabilistic networks from data.
\newblock {\em Machine learning}, 9(4):309--347, 1992.

\bibitem{friedman2003being}
Nir Friedman and Daphne Koller.
\newblock Being bayesian about network structure. a bayesian approach to structure discovery in bayesian networks.
\newblock {\em Machine learning}, 50:95--125, 2003.

\bibitem{gasse2014hybrid}
Maxime Gasse, Alex Aussem, and Haytham Elghazel.
\newblock A hybrid algorithm for bayesian network structure learning with application to multi-label learning.
\newblock {\em Expert Systems with Applications}, 41(15):6755--6772, 2014.

\bibitem{heckerman1995learning}
David Heckerman, Dan Geiger, and David~M Chickering.
\newblock Learning bayesian networks: The combination of knowledge and statistical data.
\newblock {\em Machine learning}, 20(3):197--243, 1995.

\bibitem{herskovits1990kutato}
E~Herskovits.
\newblock Kutato: An entropy-driven system for construction of probabilistic expert systems from databases.
\newblock In {\em Proc. 6th International Conference on Uncertainty in Artificial Intelligence, Cambridge, MA, 1990}, pages 117--128, 1990.

\bibitem{kitson2024eliminating}
Neville~K Kitson and Anthony~C Constantinou.
\newblock Eliminating variable order instability in greedy score-based structure learning.
\newblock In {\em International Conference on Probabilistic Graphical Models}, pages 147--163. PMLR, 2024.

\bibitem{kitson2024impact}
Neville~K Kitson and Anthony~C Constantinou.
\newblock The impact of variable ordering on bayesian network structure learning.
\newblock {\em Data Mining and Knowledge Discovery}, pages 1--25, 2024.

\bibitem{kitson2023survey}
Neville~Kenneth Kitson, Anthony~C Constantinou, Zhigao Guo, Yang Liu, and Kiattikun Chobtham.
\newblock A survey of bayesian network structure learning.
\newblock {\em Artificial Intelligence Review}, pages 1--94, 2023.

\bibitem{koller2009probabilistic}
Daphne Koller and Nir Friedman.
\newblock {\em Probabilistic graphical models: principles and techniques}.
\newblock MIT press, 2009.

\bibitem{kyrimi2021comprehensive}
Evangelia Kyrimi, Scott McLachlan, Kudakwashe Dube, Mariana~R Neves, Ali Fahmi, and Norman Fenton.
\newblock A comprehensive scoping review of bayesian networks in healthcare: Past, present and future.
\newblock {\em Artificial Intelligence in Medicine}, 117:102108, 2021.

\bibitem{larranaga1996learning}
Pedro Larranaga, Cindy~MH Kuijpers, Roberto~H Murga, and Yosu Yurramendi.
\newblock Learning bayesian network structures by searching for the best ordering with genetic algorithms.
\newblock {\em IEEE transactions on systems, man, and cybernetics-part A: systems and humans}, 26(4):487--493, 1996.

\bibitem{leonelli2024bnrep}
Manuele Leonelli.
\newblock bnrep: A repository of bayesian networks from the academic literature.
\newblock {\em arXiv preprint arXiv:2409.19158}, 2024.

\bibitem{margaritis1999bayesian}
Dimitris Margaritis and Sebastian Thrun.
\newblock Bayesian network induction via local neighborhoods.
\newblock {\em Advances in neural information processing systems}, 12, 1999.

\bibitem{pearl2009causality}
J.~Pearl.
\newblock {\em Causality}.
\newblock Causality: Models, Reasoning, and Inference. Cambridge University Press, 2009.

\bibitem{pearl1988probabilistic}
Judea Pearl.
\newblock {\em Probabilistic reasoning in intelligent systems: networks of plausible inference}.
\newblock Morgan kaufmann, 1988.

\bibitem{peters2014causal}
Jonas Peters, Joris~M Mooij, Dominik Janzing, and Bernhard Sch{\"o}lkopf.
\newblock Causal discovery with continuous additive noise models.
\newblock {\em The Journal of Machine Learning Research}, 15(1):2009--2053, 2014.

\bibitem{ramsey2017million}
Joseph Ramsey, Madelyn Glymour, Ruben Sanchez-Romero, and Clark Glymour.
\newblock A million variables and more: the fast greedy equivalence search algorithm for learning high-dimensional graphical causal models, with an application to functional magnetic resonance images.
\newblock {\em International journal of data science and analytics}, 3:121--129, 2017.

\bibitem{ramsey2018tetrad}
Joseph~D Ramsey, Kun Zhang, Madelyn Glymour, Ruben~Sanchez Romero, Biwei Huang, Imme Ebert-Uphoff, Savini Samarasinghe, Elizabeth~A Barnes, and Clark Glymour.
\newblock Tetrad—a toolbox for causal discovery.
\newblock In {\em 8th international workshop on climate informatics}, pages 1--4, 2018.

\bibitem{bnrepository}
Marco Scutari.
\newblock {Bayesian Network Repository}, 2021.
\newblock {https://www.bnlearn.com/bnrepository/}.

\bibitem{bnlearn}
Marco Scutari.
\newblock {bnlearn (Version 4.7) [Computer program]}, 2021.
\newblock {https://cran.r-project.org/web/packages/bnlearn/index.html} (downloaded: 17 December 2021).

\bibitem{scutari2019learns}
Marco Scutari, Catharina~Elisabeth Graafland, and Jos{\'e}~Manuel Guti{\'e}rrez.
\newblock Who learns better bayesian network structures: Accuracy and speed of structure learning algorithms.
\newblock {\em International Journal of Approximate Reasoning}, 115:235--253, 2019.

\bibitem{spirtes1991algorithm}
Peter Spirtes and Clark Glymour.
\newblock An algorithm for fast recovery of sparse causal graphs.
\newblock {\em Social science computer review}, 9(1):62--72, 1991.

\bibitem{spirtes2001causation}
Peter Spirtes, Clark Glymour, and Richard Scheines.
\newblock {\em Causation, prediction, and search}.
\newblock MIT press, 2001.

\bibitem{suzuki1999learning}
Joe Suzuki.
\newblock Learning bayesian belief networks based on the minimum description length principle: basic properties.
\newblock {\em IEICE transactions on fundamentals of electronics, communications and computer sciences}, 82(10):2237--2245, 1999.

\bibitem{tsamardinos2003time}
Ioannis Tsamardinos, Constantin~F Aliferis, and Alexander Statnikov.
\newblock Time and sample efficient discovery of markov blankets and direct causal relations.
\newblock In {\em Proceedings of the ninth ACM SIGKDD international conference on Knowledge discovery and data mining}, pages 673--678, 2003.

\bibitem{tsamardinos2006max}
Ioannis Tsamardinos, Laura~E Brown, and Constantin~F Aliferis.
\newblock The max-min hill-climbing bayesian network structure learning algorithm.
\newblock {\em Machine learning}, 65(1):31--78, 2006.

\bibitem{verma1990equivalence}
Thomas Verma and Judea Pearl.
\newblock Equivalence and synthesis of causal models.
\newblock In {\em Proceedings of the Sixth Annual Conference on Uncertainty in Artificial Intelligence}, pages 255--270, 1990.

\bibitem{vitolo2018modeling}
Claudia Vitolo, Marco Scutari, Mohamed Ghalaieny, Allan Tucker, and Andrew Russell.
\newblock Modeling air pollution, climate, and health data using bayesian networks: A case study of the english regions.
\newblock {\em Earth and Space Science}, 5(4):76--88, 2018.

\bibitem{yuan2011learning}
Changhe Yuan, Brandon Malone, and Xiaojian Wu.
\newblock Learning optimal bayesian networks using a* search.
\newblock In {\em Twenty-second international joint conference on artificial intelligence}, pages 2186--2191, 2011.

\bibitem{zheng2018dags}
Xun Zheng, Bryon Aragam, Pradeep~K Ravikumar, and Eric~P Xing.
\newblock Dags with no tears: Continuous optimization for structure learning.
\newblock {\em Advances in neural information processing systems}, 31, 2018.

\end{thebibliography}

\end{document}